\documentclass{article}

\usepackage{tablefootnote}
\usepackage{placeins}
\usepackage{arxiv}
\usepackage{subcaption}
\usepackage[utf8]{inputenc} 
\usepackage[T1]{fontenc}    
\usepackage{hyperref}       
\usepackage{url}            
\usepackage{booktabs}       
\usepackage{amsfonts}       
\usepackage{nicefrac}       
\usepackage{microtype}      
\usepackage{lipsum}		
\usepackage{graphicx}
\usepackage{natbib}
\usepackage{doi}
\usepackage[dvipsnames]{xcolor}
\definecolor{mypink1}{rgb}{0.858, 0.188, 0.478}

\title{Progressive Knowledge Distillation of \emph{Stable Diffusion XL} using Layer Level Loss}

\date{} 					

\author{ {\hspace{1mm}Yatharth Gupta}\thanks{Equal Contribution} \\
	Segmind\\
	\texttt{yatharthg@segmind.com} \\
	\And
	{\hspace{1mm}Vishnu V.~Jaddipal*} \\
	Segmind\\
	\texttt{vishnuj@segmind.com} \\
	\AND
	Harish Prabhala
         \\
	Segmind\\
	\texttt{harish@segmind.com} \\
 	\And
	Sayak Paul\\
	Hugging Face\\
	\texttt{sayak@huggingface.co} \\
 	\And
	Patrick Von Platen\\
	Hugging Face\\
	\texttt{patrick@huggingface.co} \\
}

\renewcommand{\headeright}{}
\renewcommand{\undertitle}{Technical Report}

\hypersetup{
pdftitle={Progressive Knowledge Distillation Of Stable Diffusion XL Using Layer Level Loss},
pdfsubject={},
pdfauthor={Yatharth Gupta, Vishnu V. Jaddipal},
pdfkeywords={},
}

\begin{document}
\maketitle

\begin{abstract}

Stable Diffusion XL (SDXL) has become the best open source text-to-image model (T2I) for its versatility and top-notch image quality. Efficiently addressing the computational demands of SDXL models is crucial for wider reach and applicability. In this work, we introduce two scaled-down variants, Segmind Stable Diffusion (SSD-1B) and Segmind-Vega, with 1.3B and 0.74B parameter UNets, respectively, achieved through progressive removal using layer-level losses focusing on reducing the model size while preserving generative quality. We release these models weights at \textit{\textcolor{mypink1}{https://hf.co/Segmind}}.

Our methodology involves the elimination of residual networks and transformer blocks from the U-Net structure of SDXL, resulting in significant reductions in parameters, and latency. Our compact models effectively emulate the original SDXL by capitalizing on transferred knowledge, achieving competitive results against larger multi-billion parameter SDXL.

Our work underscores the efficacy of knowledge distillation coupled with layer-level losses in reducing model size while preserving the high-quality generative capabilities of SDXL, thus facilitating more accessible deployment in resource-constrained environments.
\end{abstract}

\section{Introduction}

Stable Diffusion~\citep{sd} has emerged as highly influential in the realm of text-to-image (T2I) synthesis, playing a pivotal role as an open-source framework. Its remarkable capabilities has spurred its integration as a backbone in various text-guided vision applications. Stable Diffusion, characterized as T2I-specialized latent diffusion models (LDMs), leverages diffusion operations within a semantically compressed space, enhancing computational efficiency. Central to the architecture of Stable Diffusion is a U-Net that employs iterative sampling to progressively denoise a random latent code. This process is further supported by a text encoder and an image decoder, orchestrating the generation of text-aligned images. SDXL~\citep{sdxl} is the largest variant with a 2.6B Parameter UNet and two text encoders, providing the best quality among open source models.

Notably, distillation techniques have been applied to pretrained diffusion models to curtail the number of denoising steps, resulting in identically structured models with reduced sampling requirements. Additionally, methods such as post-training quantization and implementation optimizations have been explored. The exploration of removing architectural elements in large diffusion models has also been investigated for the base U-Net models ~\citep{bksdm}. In this context, our work endeavors to apply knowledge distillation methods to the SDXL model~\citep{sdxl}, resulting in the creation of two streamlined variants, namely Segmind Stable Diffusion (SSD-1B) and Segmind-Vega. We use the base model as well as finetuned versions in the distillation process. These models, with 1.3B and 0.74B parameter UNets respectively, employ layer level losses to progressively reduce the model size to 20\%, 40\%, 50\%, 60\%, and ultimately 70\%. This reduction in model size aims to strike a balance between computational efficiency and the preservation of generative capabilities, making SDXL more accessible for diverse applications.

\section{Related Work}

\subsection{Large Latent Diffusion Models}
The exploration of diffusion-based generative models has been instrumental in achieving high-fidelity synthesis with broad mode coverage by gradually removing noise from corrupted data. The integration of these models with pretrained language models has notably enhanced the quality of text-to-image (T2I) synthesis. In models such as Imagen ~\citep{imagen} and Deepfloyd IF~\citep{deepfloyd}, text-conditional diffusion models generate small images, subsequently upsampled through super-resolution modules. DALL·E~\citep{dalle} style models, on the other hand, employ a text-conditional prior network to produce an image embedding, transformed via a diffusion decoder and further upscaled into higher resolutions. LDMs perform diffusion modeling in a low-dimensional latent space constructed through a pixel-space autoencoder. 

\subsection{Efficient Diffusion Models}
Efforts to address the slow sampling process in diffusion models have been widespread. Diffusion-tailored distillation progressively transfers knowledge from a pretrained diffusion model to a model with fewer sampling steps while maintaining the same architecture. Latent Consistency Models ~\citep{lcm} also allow the models to generate images in very few steps. Combining this with Low Rank Adapters (LoRAs)~\citep{lcmlora} provides a very easy way of enabling fast generation with large models. Fast high-order solvers for diffusion ordinary differential equations aim to boost sampling speed. In complement to these approaches, our network compression method reduces per-step computation and seamlessly integrates with models employing fewer sampling steps. Leveraging quantization and implementation optimizations designed for SDXL can further enhance the efficiency of our compact models.

\subsection{Distillation-Based Compression}
Knowledge Distillation (KD) has been successful in improving the performance of small-size models by exploiting output-level and feature-level information from larger source models. While classical KD has found applications in efficient GANs, and Stable Diffusion Base model. Our work demonstrates the extension of distillation pretraining techniques, proven successful in small yet capable general-purpose language models and vision transformers, to SDXL.

\subsubsection{Concurrent Studies}
Studies such as SnapFusion \citep{snapfusion} achieve an efficient U-Net for Stable Diffusion through architecture evolution and step distillation. Wuerstchen ~\citep{wuerstchen} introduces two diffusion processes on low- and high-resolution latent spaces for economic training. While these works are valuable, it is essential to note that they often require significantly larger computational resources than our proposed approach. Additionally, As demonstrated on Stable Diffusion, BK-SDM proposes pruning the UNet via removal of blocks, showcasing promising compression.

This work uses the technique of classical architectural compression in achieving smaller and faster diffusion models. The approach involves the removal of multiple transformer layers from the U-Net of SDXL, followed by retraining with feature-level knowledge distillation for general-purpose T2I. The contributions of this study are summarized as follows:
\begin{itemize}

\item Architectural Compression: We compress SDXL by strategically removing architectural blocks from the U-Net, resulting in a notable reduction in model size (up to 70\%) and increased inference speeds(up to 100\% speedup). 

\item Feature Distillation: We use feature distillation for training diffusion models, demonstrating its remarkable benefits in achieving competitive T2I performance with significantly fewer resources. The cost-effectiveness of network compression is emphasized, particularly when compared to the substantial expense of training diffusion models from scratch.

\item Downstream benefits: The method, to an extent preserves fidelity of generation with different LoRA and Controlnet networks, thus requiring less training to be used on the distilled model.

\end{itemize}

In summary, this research explores classical architectural compression for SDXL, providing a cost-effective strategy for building compact general-purpose diffusion models with compelling performance.

\section{Methodology}
\label{sec:Methodology}

In our pursuit of compressing SDXL models, we adopt a nuanced approach that centers on the removal of transformer layers within attention blocks. Our observation reveals a redundancy in numerous blocks, and our strategy involves judicious elimination without compromising the model's generative prowess. We draw inspiration from the architectural compression techniques applied to Stable Diffusion v1.5's \footnote{\url{https://huggingface.co/runwayml/stable-diffusion-v1-5}} U-Net and extend the methodology to SDXL, yielding two scaled-down variants: Segmind Stable Diffusion (SSD-1B) and Segmind-Vega.

\subsection{Architecture}
Our compression strategy is motivated by the recognition that certain layers are dispensable without significantly affecting the model's performance. We leverage insights from various teacher models, including SDXL-base-1.0 and the fine-tuned Zavychroma-XL \footnote{\url{https://civitai.com/models/119229/zavychromaxl}} and Juggernaut-XL \footnote{\url{https://civitai.com/models/133005?modelVersionId=240840}}, during the compression process.

We report similar findings as BK-SDM ~\citep{bksdm}, in that the middle block of the U-Net can be removed without significantly affecting image quality. To add, we observe that removal of only the attention layers and the second residual network \citep{resnet} block preserves image quality to a higher degree, as opposed to removal of the whole mid-block.
\FloatBarrier
\begin{figure}[!h]
    \centering
    \includegraphics[width=1\linewidth]{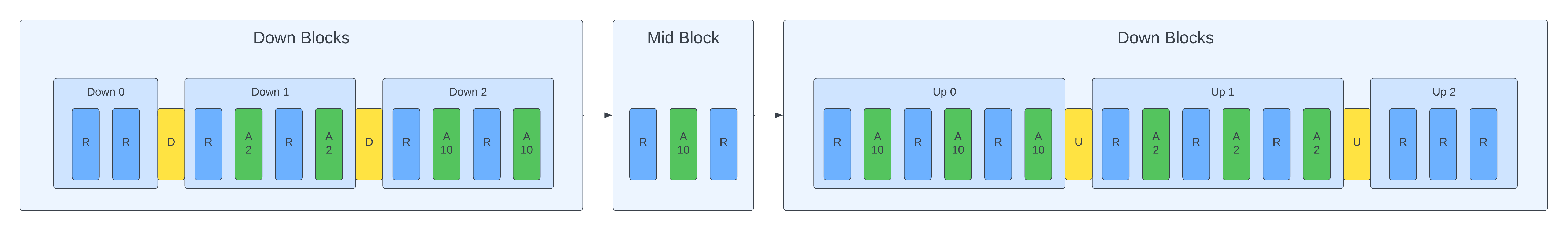}
    \caption{SDXL U-Net structure}
    \label{fig:sdxlarch}
\end{figure}
\begin{figure}[!h]
    \centering
    \includegraphics[width=1\linewidth]{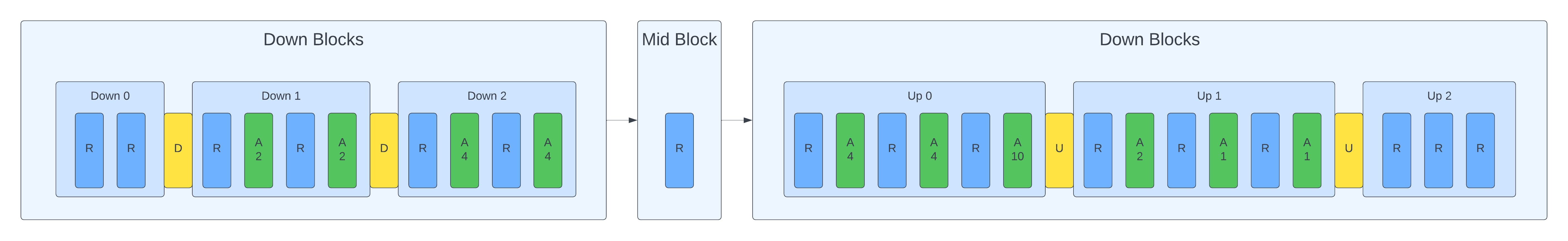}
    \caption{SSD-1B U-Net structure}
    \label{fig:ssdarch}
\end{figure}
\begin{figure}[!h]
    \centering
    \includegraphics[width=1\linewidth]{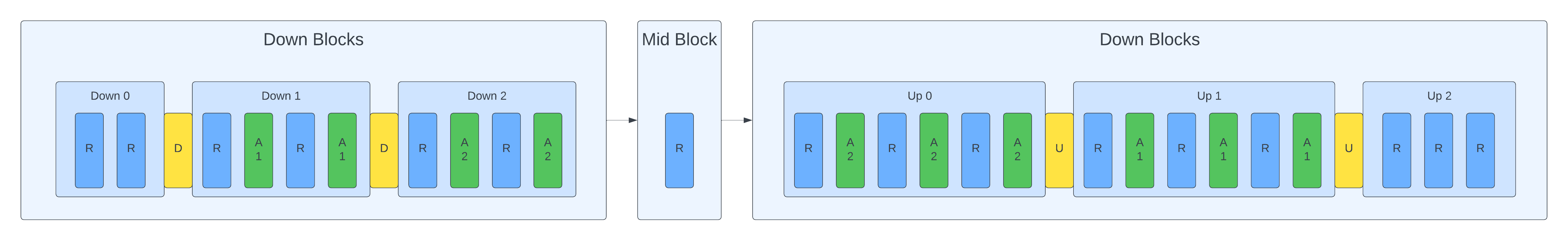}
    \caption{Vega U-Net structure}
    \label{fig:vegaarch}
\end{figure}
\FloatBarrier
\subsection{Loss}
In contrast to the block-level losses employed in prior work, we introduce layer-level losses specific to each attention and ResNet layer. This refined approach allows for a more granular assessment of the model's internal representations, enabling us to identify and retain essential features while discarding redundant elements. Our choice of layer-level losses is influenced by their efficacy in capturing the nuanced interactions within the model's architecture.

\subsubsection{Input Generation}

To obtain the input for the U-Net, we employ pretrained text encoders for the to obtain the text embeddings. The latent representations of the image are obtained by the pretrained VAE. Both text encoders and the VAE are kept frozen during training and only the UNet is trained. The latent representation
\emph{z} of an image and its paired text embedding \emph{y} form the
basis for our training process.

\subsubsection{Task Loss}

We formulate the task loss, denoted as \emph{$L_{Task}$}, which is computed through the reverse denoising process. The task loss measures the disparity between the sampled noise $\epsilon$ from the diffusion process and the estimated noise $\epsilon_S(z_t, y, t)$ generated by our compact UNet student. The objective is to align the noise distribution of the student with that of the teacher.

\begin{center}
$L_{Task} = \mathbb{E}_{z,\epsilon,y,t,t_h} \textbar\textbar{}\epsilon - \epsilon_S (z_t, y, t)\textbar\textbar_2^2$
\end{center}

\subsubsection{Output-Level Knowledge Distillation (KD)}

The compact student is trained to imitate the outputs of the original
U-Net teacher, denoted as \emph{$\epsilon_T$}, using an output-level KD
objective. This objective ensures that the overall output distribution
of the student aligns with that of the teacher.

\begin{center}
    
$L_{OutKD} = \mathbb{E}_{z,\epsilon,y,t,t_h} \textbar\textbar\epsilon_T -\epsilon_S(z_t, y, t)\textbar\textbar_2^2$
\end{center}

\subsubsection{Feature-Level Knowledge Distillation (KD)}

A pivotal component of our approach is feature-level KD, providing rich
guidance for the student's training. The feature-level KD objective,
denoted as \emph{$L_{FeatKD}$}, measures the difference between the feature
maps of corresponding layers in both the teacher and student models.
Importantly, our approach eliminates the need for additional regressors
by ensuring that the dimensionality of feature maps already matches at
the end of each layer in both models.

\begin{center}
    
$L_{FeatKD} = \mathbb{E}_{h,X_l} \textbar\textbar f_l^T(z_t, y, t) - f_l^S(z_t, y, t)\textbar\textbar_2^2$
\end{center}

\subsubsection{Overall Objective}

The final objective encompasses the task loss, output-level KD, and
feature-level KD, weighted by coefficients \emph{$\lambda_{OutKD}$} and
\emph{$\lambda_{FeatKD}$}. Without loss-weight tuning, our approach demonstrates effectiveness in empirical validation.

\begin{center}
L = $L_{Task} + \lambda_{OutKD}*L_{OutKD} + \lambda_{FeatKD}*L_{FeatKD}$
\end{center}

Another advantage of this method of distillation is that LoRA weights created for the parent model tend to produce close results without retraining. This may reduce the number of training steps required to migrate models.

To expound on our compression strategy, we consider the analogy to DistilBERT \citep{distillbert}, which reduces the number of layers while initializing the compact model with original weights. Our compression methodology involves targeted removal strategies in both down and up stages.

\subsection{Teacher Models}

We initially take SDXL Base \footnote{\url{https://huggingface.co/stabilityai/stable-diffusion-xl-base-1.0}} as the teacher, but later swap it for a finetuned model, ZavychromaXL \footnote{\url{https://civitai.com/models/119229/zavychromaxl}} and finally use JuggernautXL \footnote{\url{https://civitai.com/models/133005?modelVersionId=240840}}. We find that swapping the teacher boosts the quality significantly even if the same dataset is used again. This showcases that using multiple expert models can aid in instilling new concepts as well as improving quality of the student.

Our compression methodology, inspired by proven techniques \citep{bksdm}, not only reduces model size but also ensures that essential features are retained through the careful removal of redundant blocks. The introduction of layer-level losses further refines this process, contributing to the overall efficiency and efficacy of our compressed models—SSD-1B and Segmind-Vega.
\subsection{Pruning}
We employ human evaluation of outputs along with heuristics to identify potential attention layers to remove.

To create SSD-1B, along with removal of the mid-block's attention layers and the second Residual Network, we remove the following layers of SDXL:
\begin{itemize}
    \item 4th, 5th,7th,8th,9th and 10th transformer blocks of all attention layers in the 3rd downsampling stage and the first two attention layers of the first upsampling stage of the U-Net.
    
    \item The second transformer block of the second and third attention layers of the second upsampling stage.
\end{itemize}
To create Segmind Vega, we remove the following layers:
\begin{itemize}
    \item 3rd, 4th, 5th, 6th, 7th, 8th,9th and 10th transformer blocks of
    the first attention layer of the third downsampling stage and all attention layers in the first upsampling stage of the U-Net.
    \item 2nd, 4th, 5th, 6th, 7th, 8th,9th and 10th transformer blocks of
    the second attention layer of the third downsampling stage.
    
    \item The second transformer block of all attention layers of the second downsampling and upsampling stages.

\end{itemize}

\section{Training}

In our training methodology, we adopt a distillation-based retraining
approach. We use a layer-level loss in an attempt to mimic the features at each stage of the teacher U-Net. This process is
crucial for achieving efficient knowledge transfer and preserving the
generative quality of SDXL even in significantly compressed models.

Our training strategy, inspired by distillation-based retraining,
ensures that our compressed models inherit the essential knowledge from
the teacher model, enabling them to efficiently mimic the behavior of
the original U-Net across various layers, including attention and residual network (ResNet)
layers.

We trained SSD-1B at fp16 mixed-precision for a total of 251,000 steps with a constant learning rate of 1e-5, using Adam Optimizer \citep{adam}, at 1024*1024 image resolutions, on four 80GB A100 GPUs at an effective batch size of 32.
We trained Vega at fp16 mixed-precision for a total of 540,000 steps with a learning rate of 1e-5, at 1024*1024 image resolutions, on four 80GB A100 GPUs, at an effective batch size of 128. The datasets used for training and evaluation include GRIT \citep{grit} and images generated by Midjourney \footnote{\url{https://www.midjourney.com/}}.

\section{Results}

\begin{figure}[h]
\begin{subfigure}{0.33\textwidth}
\includegraphics[width=0.9\linewidth, height=4cm]{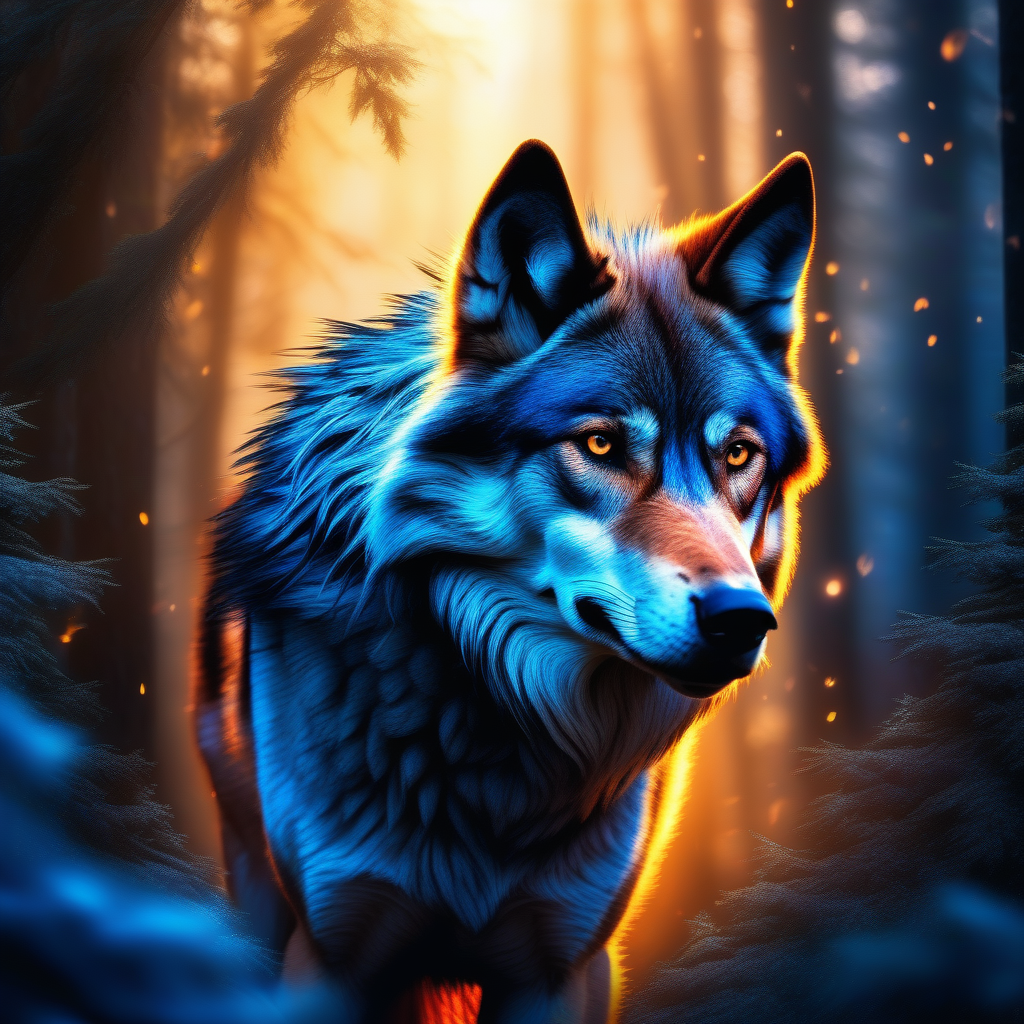} 
\caption{SDXL}
\label{fig:2subim1}
\end{subfigure}
\begin{subfigure}{0.33\textwidth}
\includegraphics[width=0.9\linewidth, height=4cm]{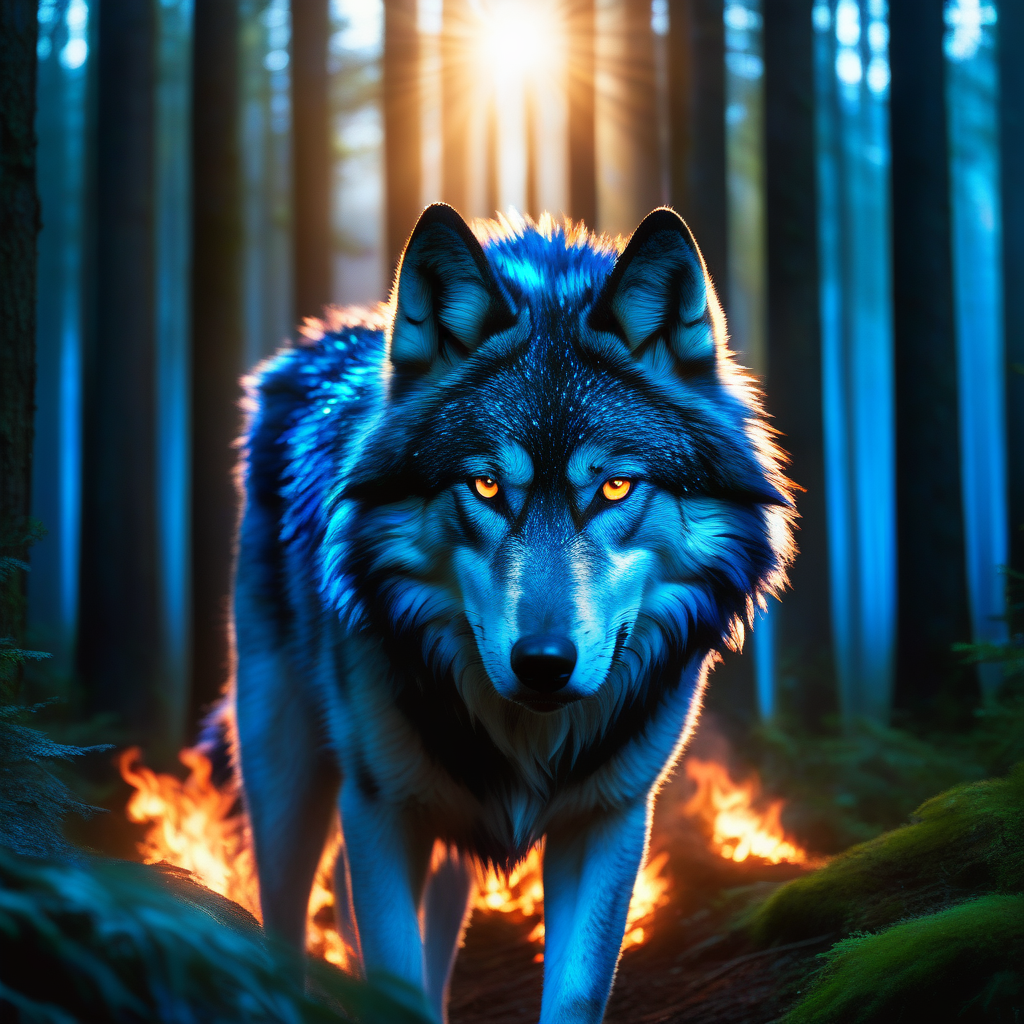}
\caption{SSD-1B}
\label{fig:1subim2}
\end{subfigure}
\begin{subfigure}{0.33\textwidth}
\includegraphics[width=0.9\linewidth, height=4cm]{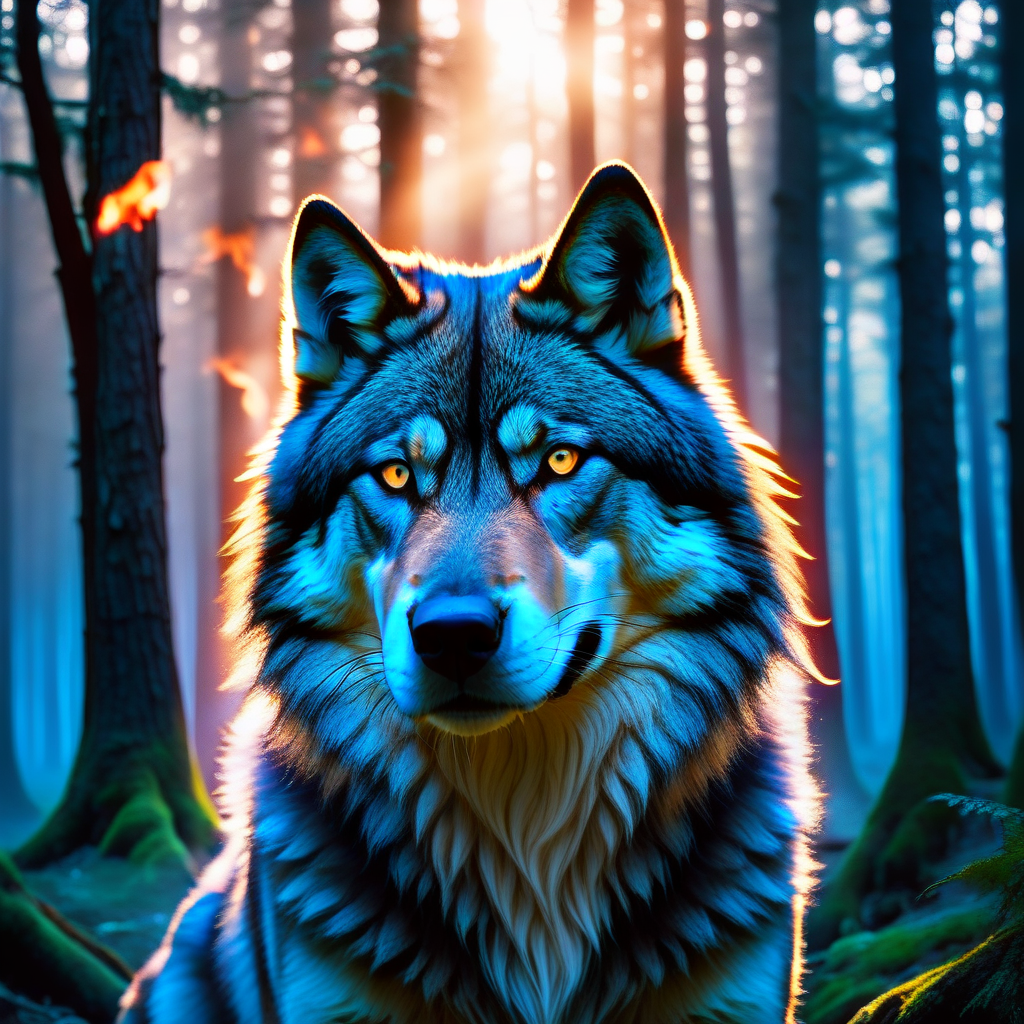}
\caption{Vega}
\label{fig:1subim3}
\end{subfigure}
\caption{"A royal flaming wolf emerging from a magical big forest, blue flames, front facing, portrait, closeup, dark, bokeh, dawn, god rays, highly detailed, highres, Cinematic, Cinemascope, astonishing, epic, gorgeous, ral-fluff"}
\label{fig:image0}
\end{figure}

\begin{figure}[h]

\begin{subfigure}{0.33\textwidth}
\includegraphics[width=0.9\linewidth, height=4cm]{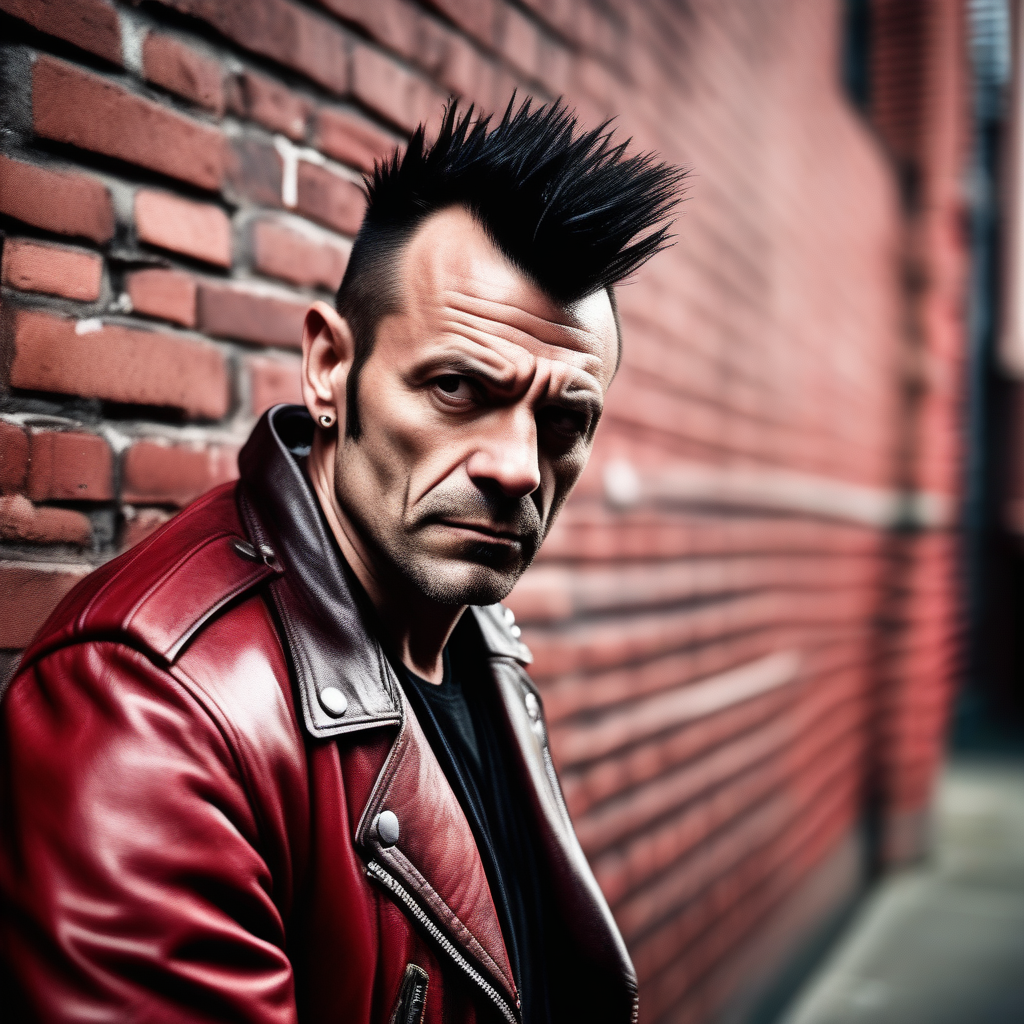} 
\caption{SDXL}
\label{fig:3subim1}
\end{subfigure}
\begin{subfigure}{0.33\textwidth}
\includegraphics[width=0.9\linewidth, height=4cm]{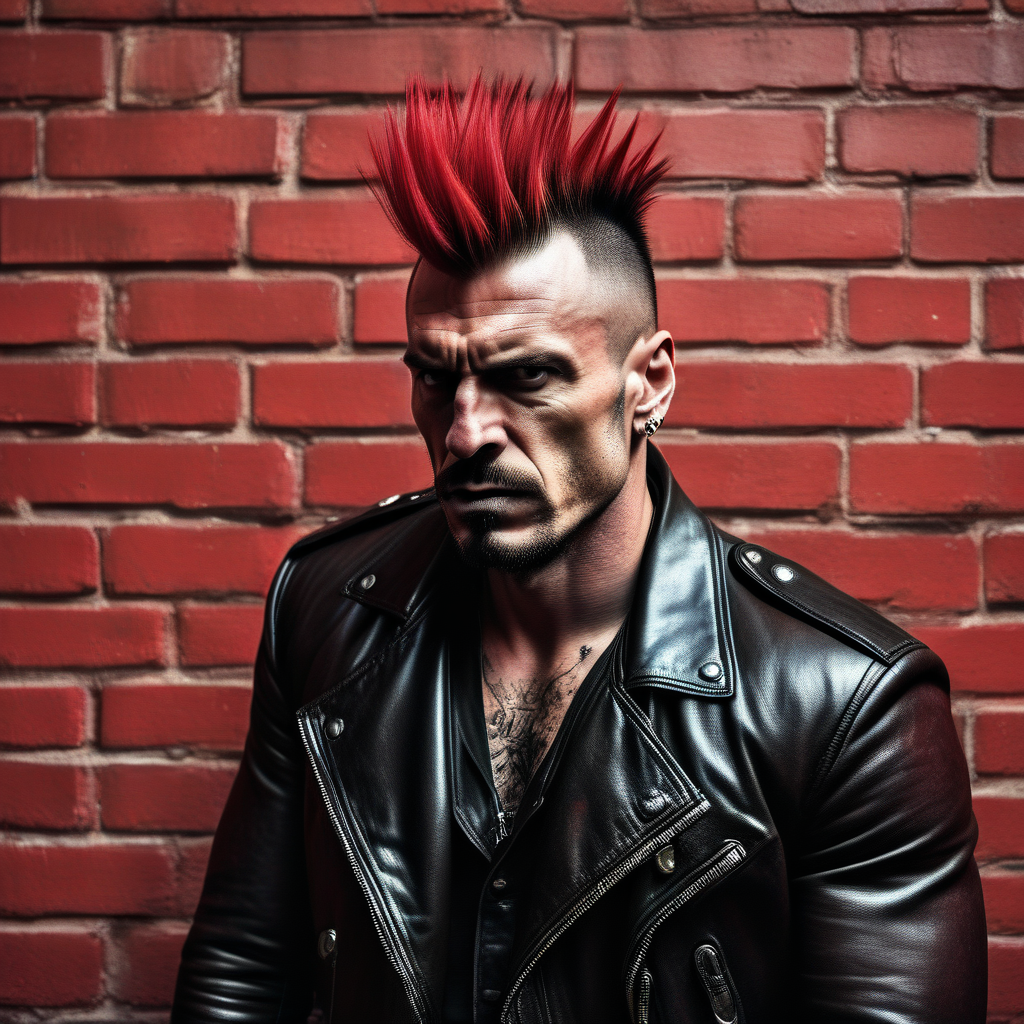}
\caption{SSD-1B}
\label{fig:2subim2}
\end{subfigure}
\begin{subfigure}{0.33\textwidth}
\includegraphics[width=0.9\linewidth, height=4cm]{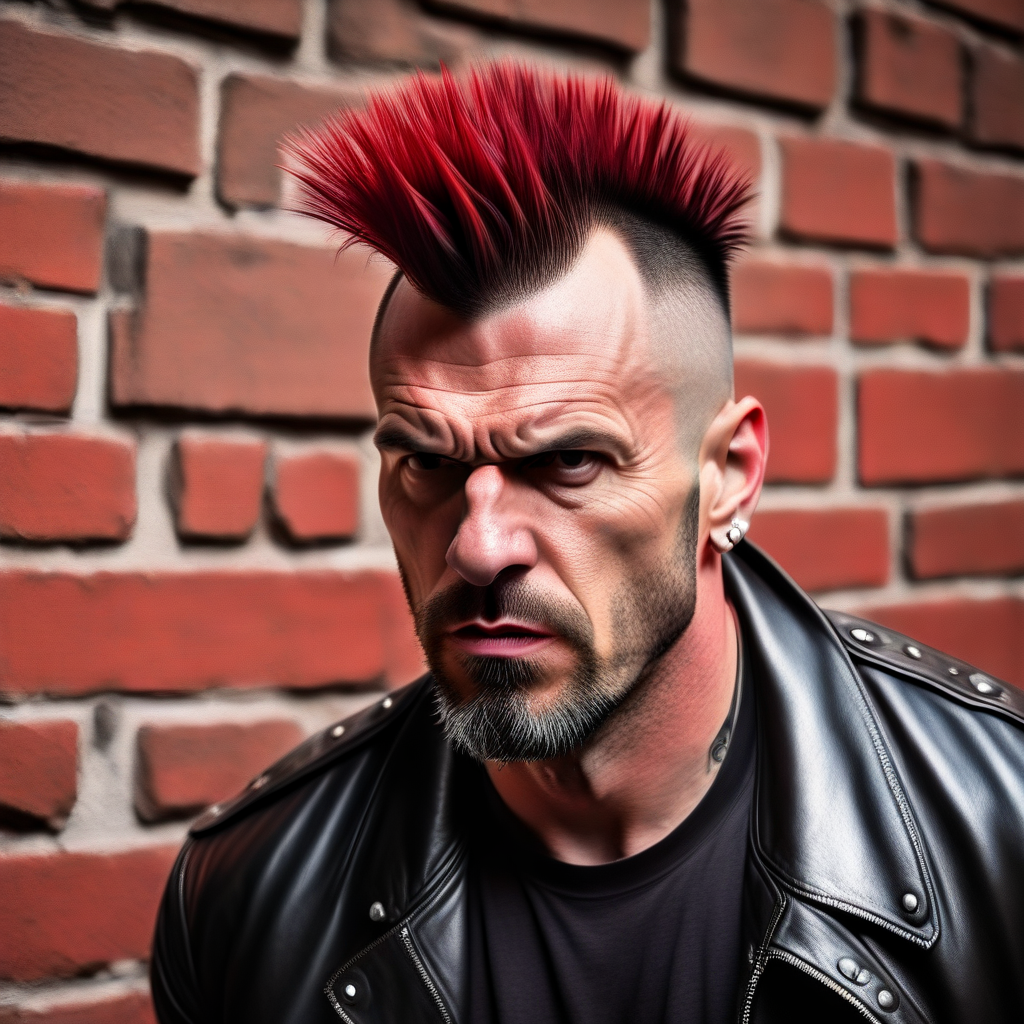}
\caption{Vega}
\label{fig:2subim3}
\end{subfigure}
\caption{"raw photo, close-up, punk band cover, red brick wall, red theme, a brutal man, 40 years old, mohawk, (manly, wide jaw:1.2), leather jacket, red shirt, (vibrant colors:0.9), film grain, bokeh, fashion magazine, hdr, highly detailed photography, (muted colors, cinematic, dim colors, soothing tones:1.2), vibrant, insanely detailed, hyperdetailed, (dark shot:1.2), (vsco:0.3), (intricate details:0.9), (hdr, hyperdetailed:1.2)"}
\label{fig:image1}
\end{figure}

\begin{figure}[h]

\begin{subfigure}{0.33\textwidth}
\includegraphics[width=0.9\linewidth, height=4cm]{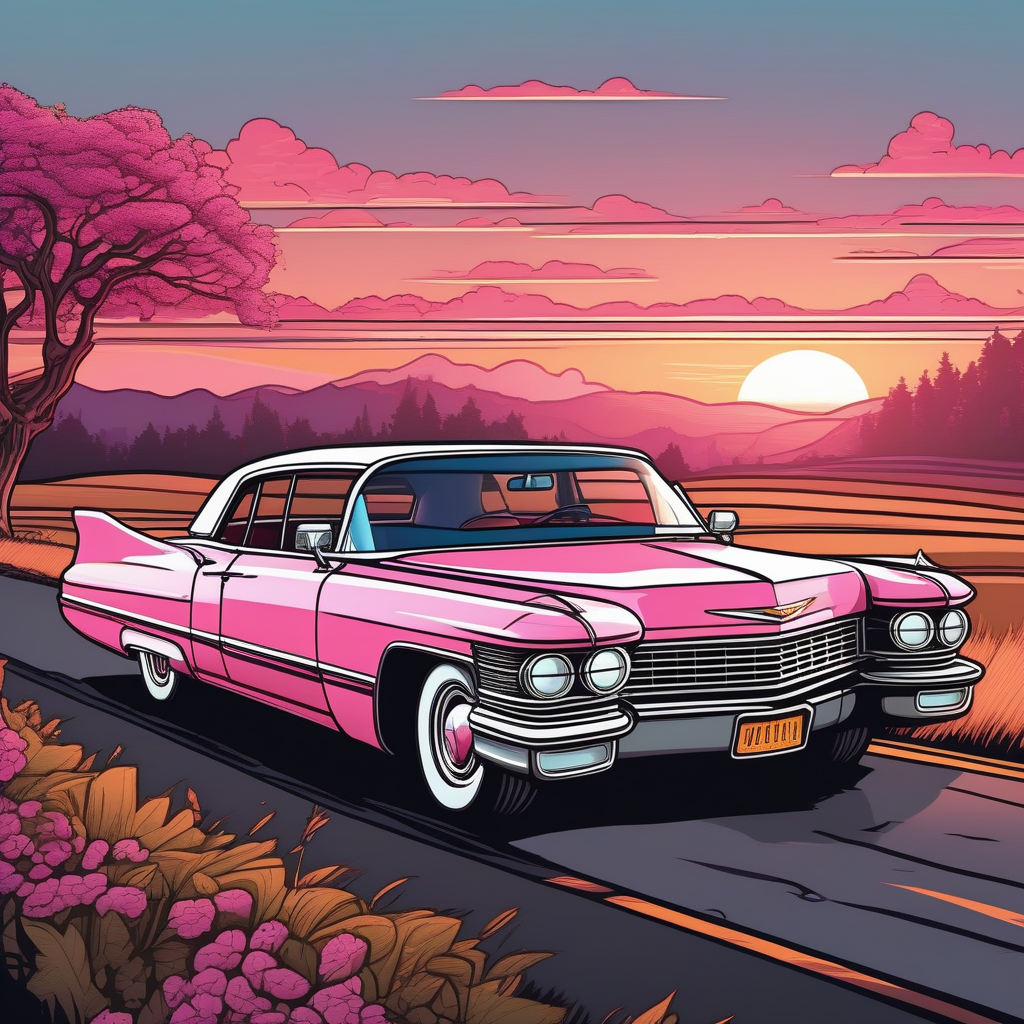} 
\caption{SDXL}
\label{fig:4subim1}
\end{subfigure}
\begin{subfigure}{0.33\textwidth}
\includegraphics[width=0.9\linewidth, height=4cm]{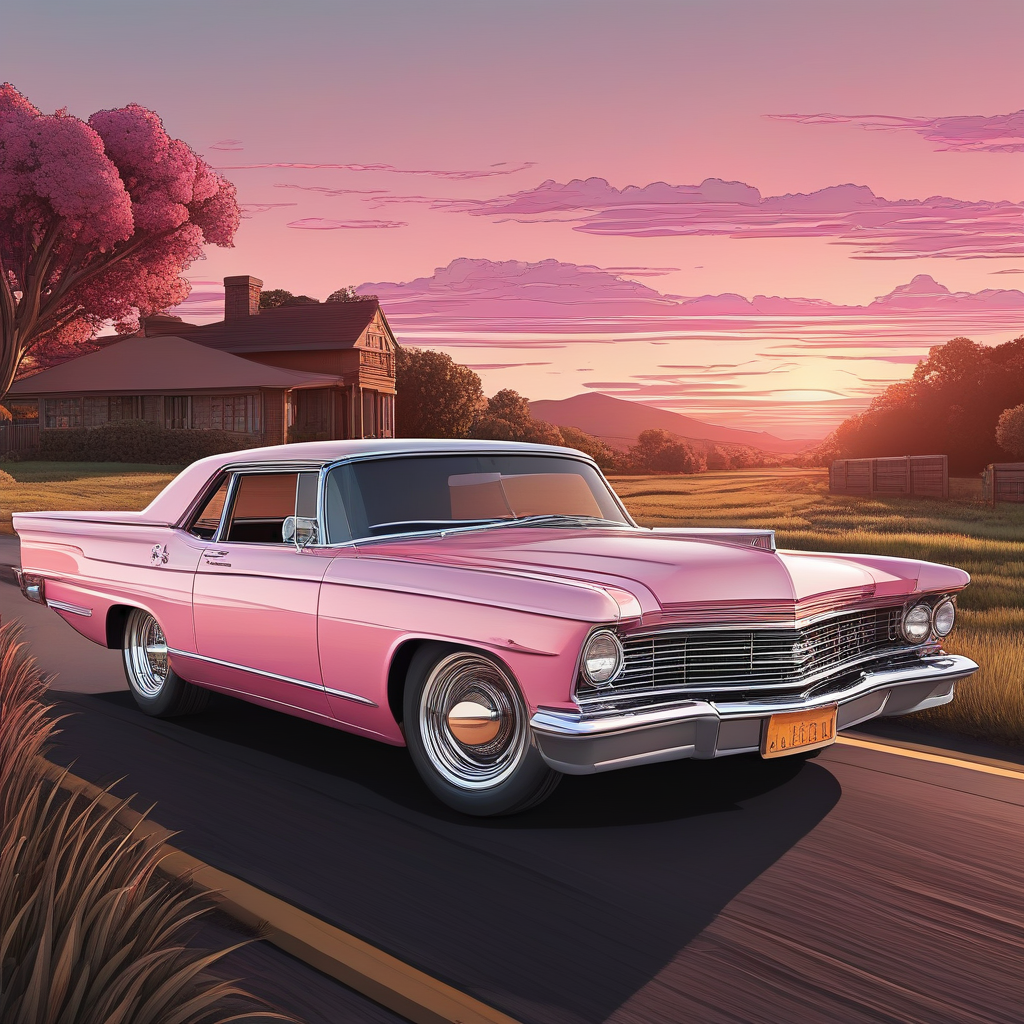}
\caption{SSD-1B}
\label{fig:3subim2}
\end{subfigure}
\begin{subfigure}{0.33\textwidth}
\includegraphics[width=0.9\linewidth, height=4cm]{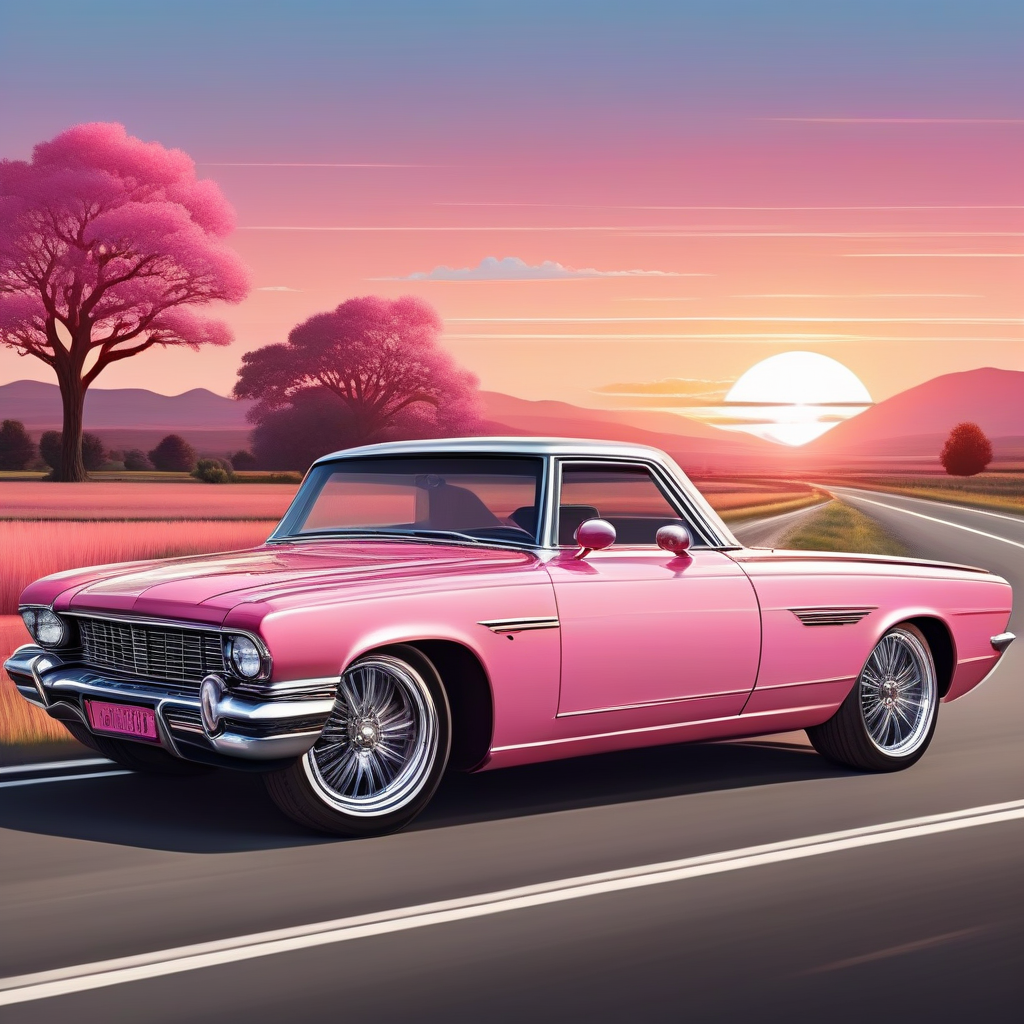}
\caption{Vega}
\label{fig:3subim3}
\end{subfigure}
\caption{"(best quality:1.5), (intricate emotional details:1.5), (sharpen details), (ultra detailed), (cinematic lighting), pink Cadillac, car, driving through the country, sunset, relaxing vibes. cartoon style, line art, sticker style"}
\label{fig:image2}
\end{figure}

\begin{figure}[h]

\begin{subfigure}{0.33\textwidth}
\includegraphics[width=0.9\linewidth, height=4cm]{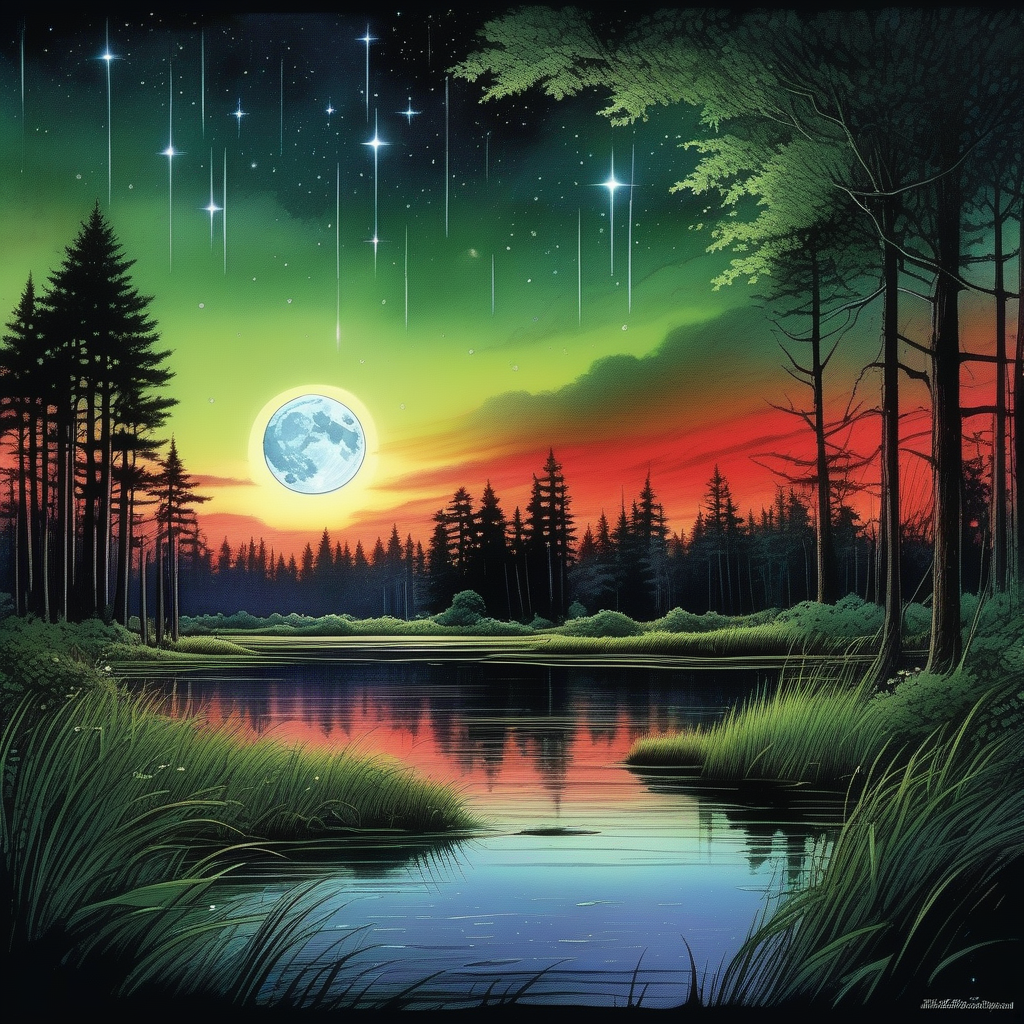} 
\caption{SDXL}
\label{fig:5subim1}
\end{subfigure}
\begin{subfigure}{0.33\textwidth}
\includegraphics[width=0.9\linewidth, height=4cm]{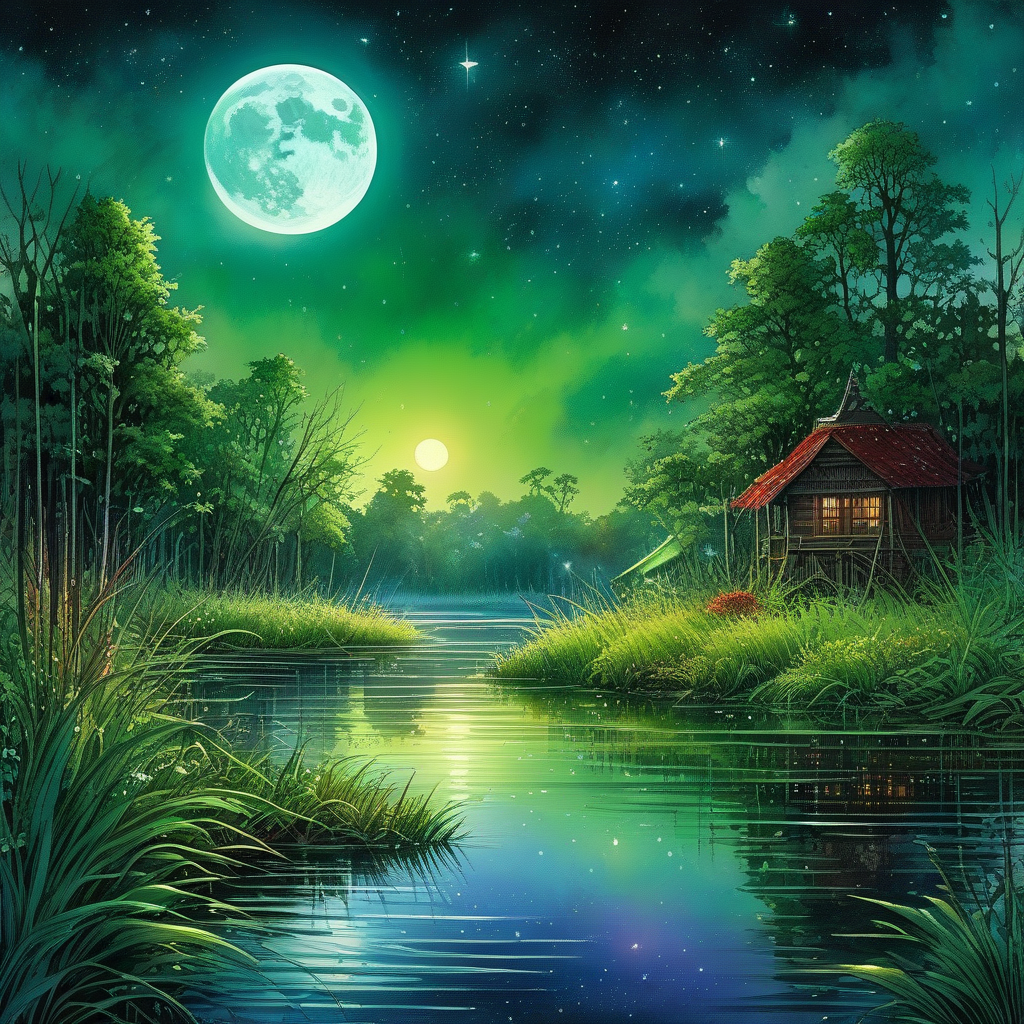}
\caption{SSD-1B}
\label{fig:4subim2}
\end{subfigure}
\begin{subfigure}{0.33\textwidth}
\includegraphics[width=0.9\linewidth, height=4cm]{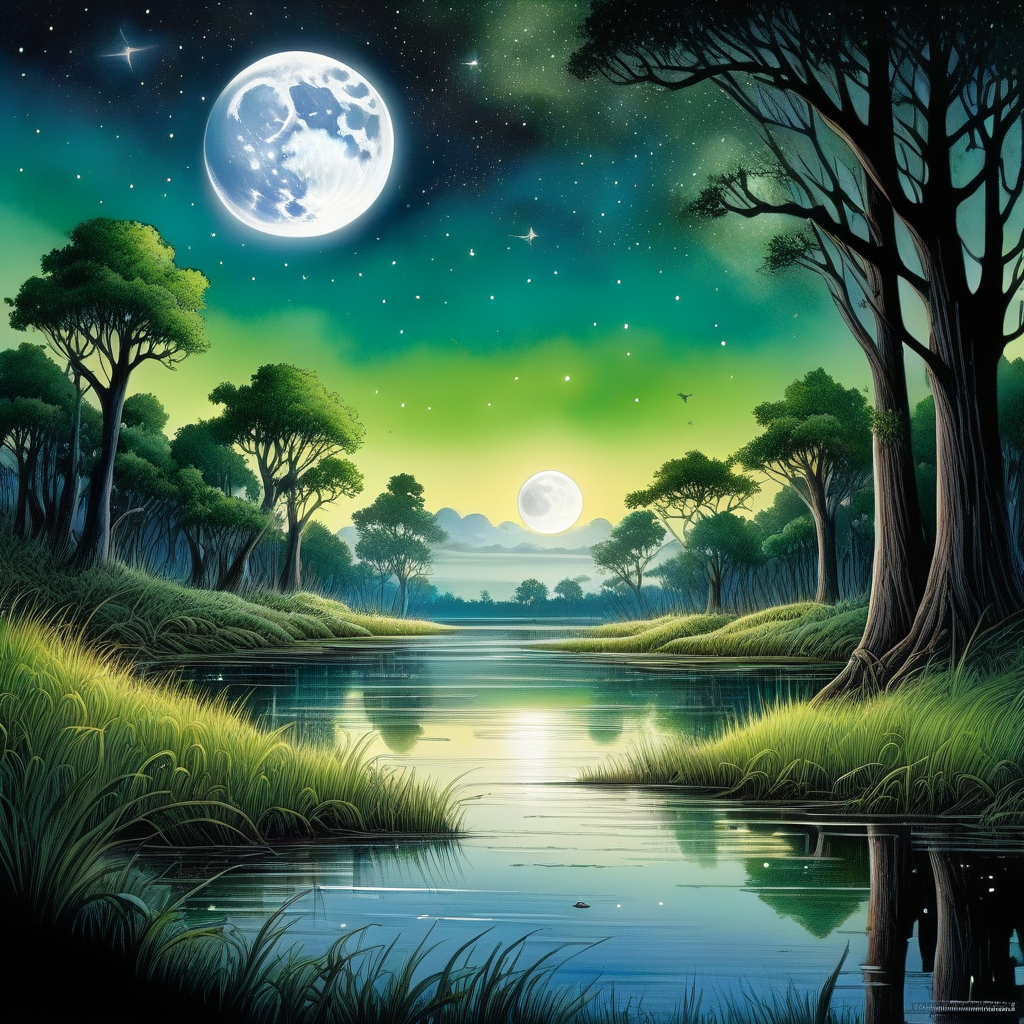}
\caption{Vega}
\label{fig:4subim3}
\end{subfigure}
\caption{"Swamp marsh Poison green red Soft watercolors digital watercolors painting illustration masterpiece raining shooting stars twinkling stars glistening stars glittery stars full moon stars full moon intricate motifs perfect composition masterpiece insanely-detailed extreme-detailed hyper-detailed beautiful volumetric deep rich colors volumetric lighting shadows Ray tracing, Mark Brooks and Dan Mumford, comic book art, perfect"}
\label{fig:image3}
\end{figure}

\begin{figure}[h]

\begin{subfigure}{0.33\textwidth}
\includegraphics[width=0.9\linewidth, height=4cm]{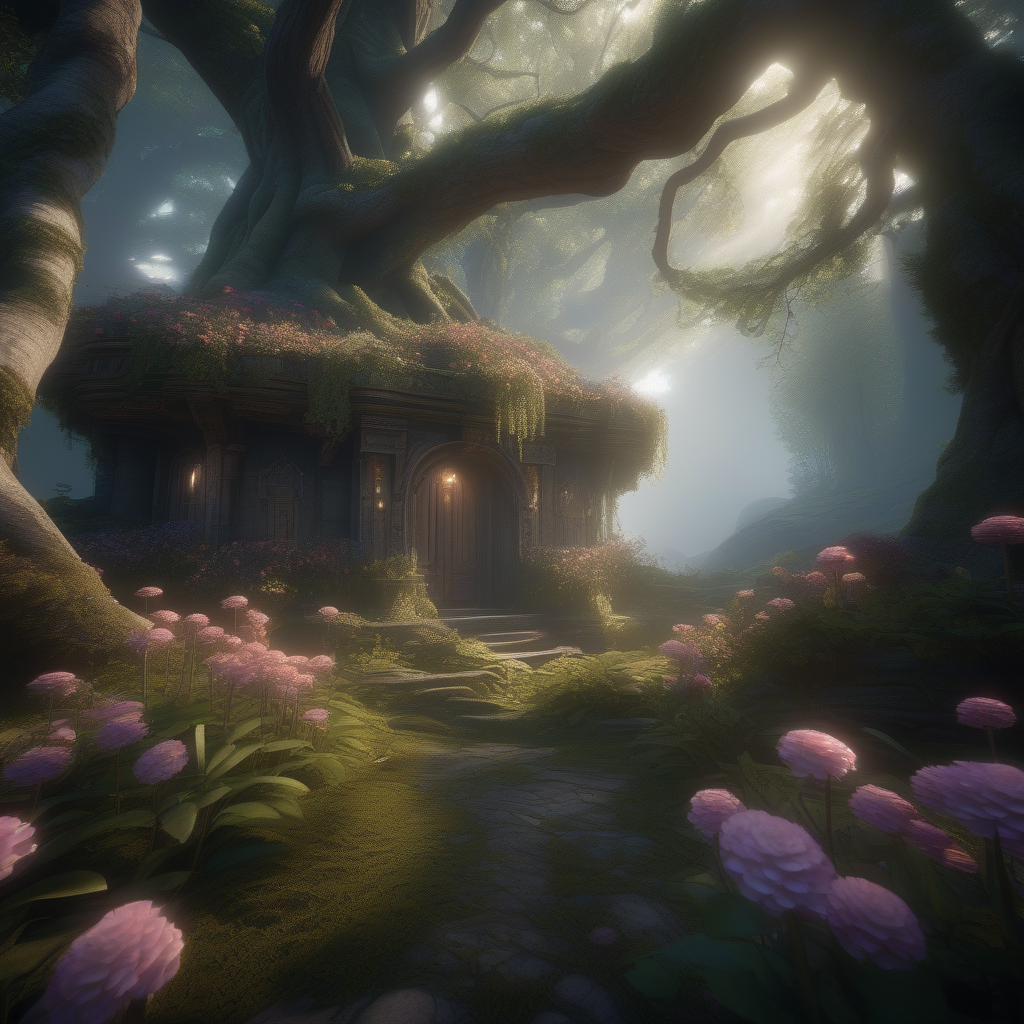} 
\caption{SDXL}
\label{fig:6subim1}
\end{subfigure}
\begin{subfigure}{0.33\textwidth}
\includegraphics[width=0.9\linewidth, height=4cm]{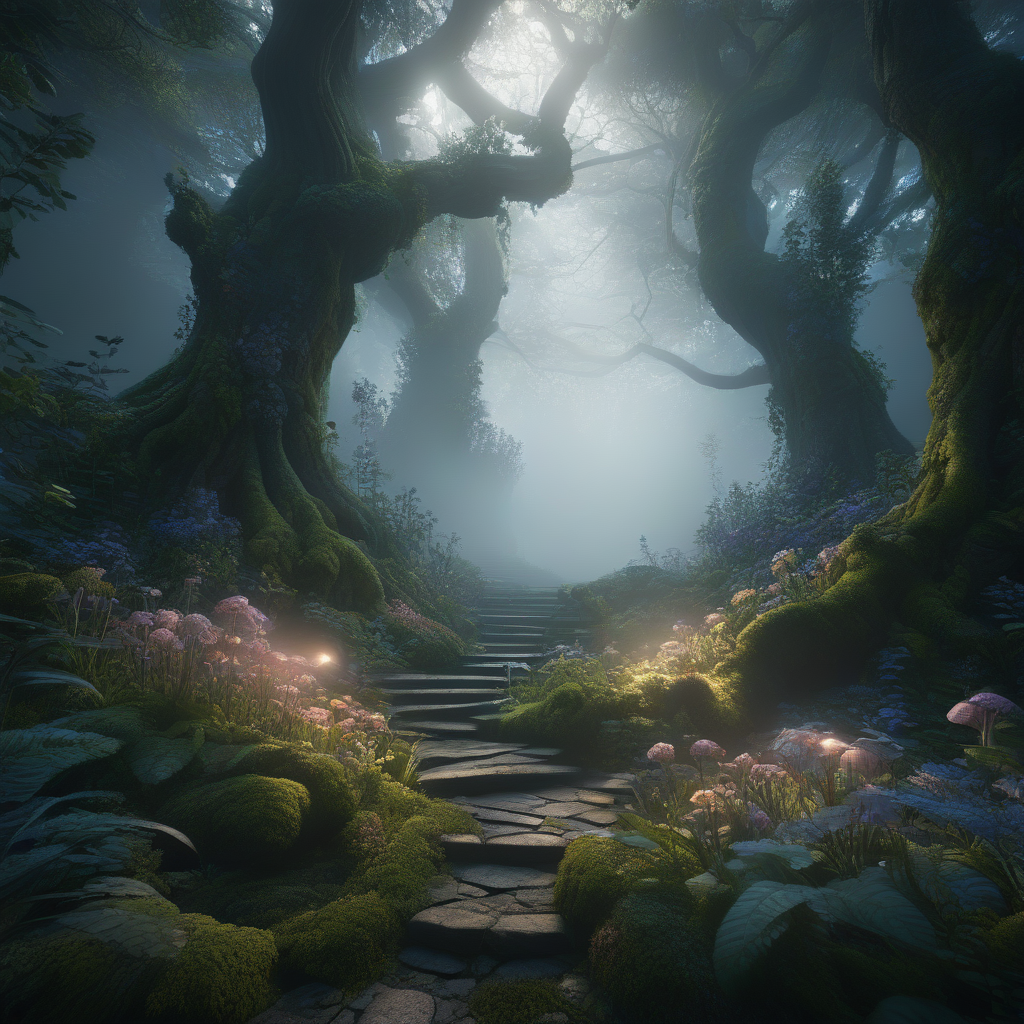}
\caption{SSD-1B}
\label{fig:5subim2}
\end{subfigure}
\begin{subfigure}{0.33\textwidth}
\includegraphics[width=0.9\linewidth, height=4cm]{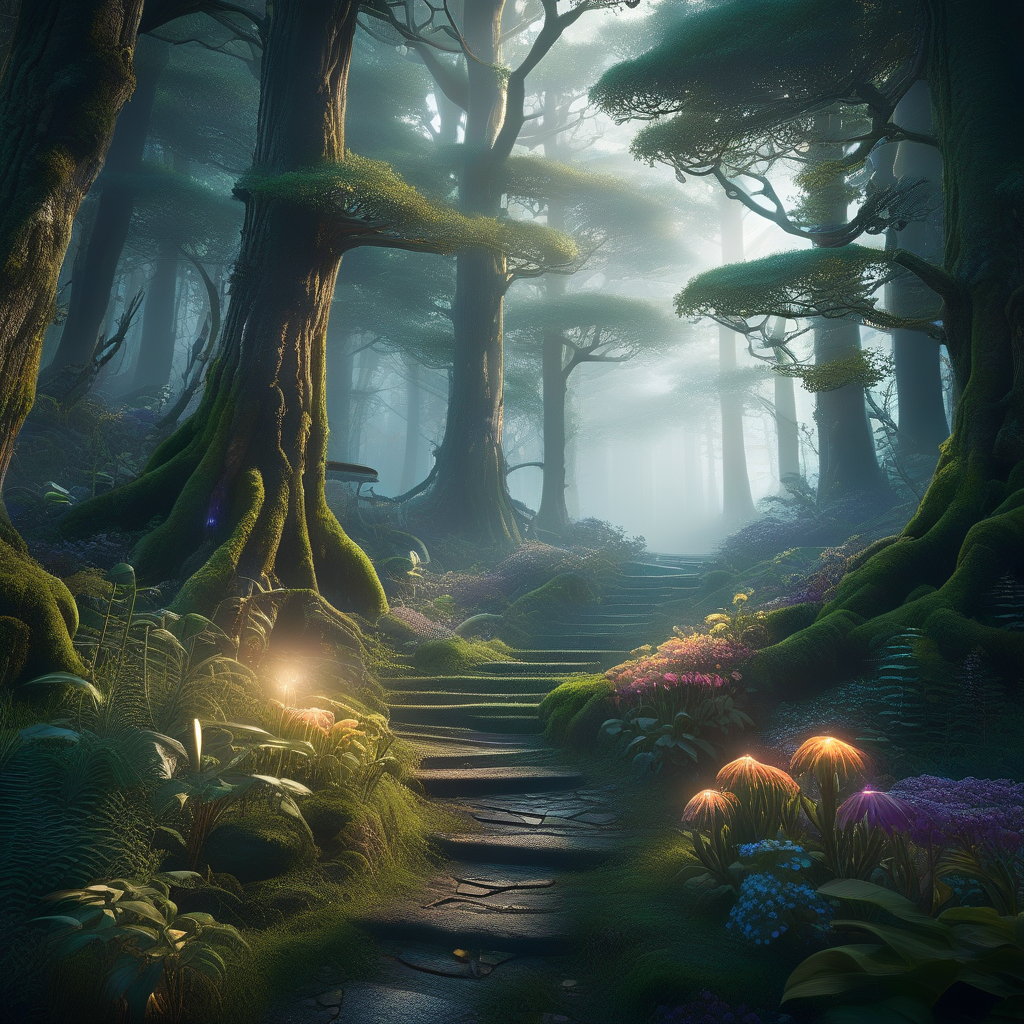}
\caption{Vega}
\label{fig:5subim3}
\end{subfigure}
\caption{"(best quality:1.5), (intricate emotional details:1.5), (sharpen details), (ultra detailed), (cinematic lighting), magical woods, unexplained lights, fantasy, otherworldy, mist, atmospheric, flowers, plants"}
\label{fig:image4}
\end{figure}

\begin{figure}[h]

\begin{subfigure}{0.33\textwidth}
\includegraphics[width=0.9\linewidth, height=4cm]{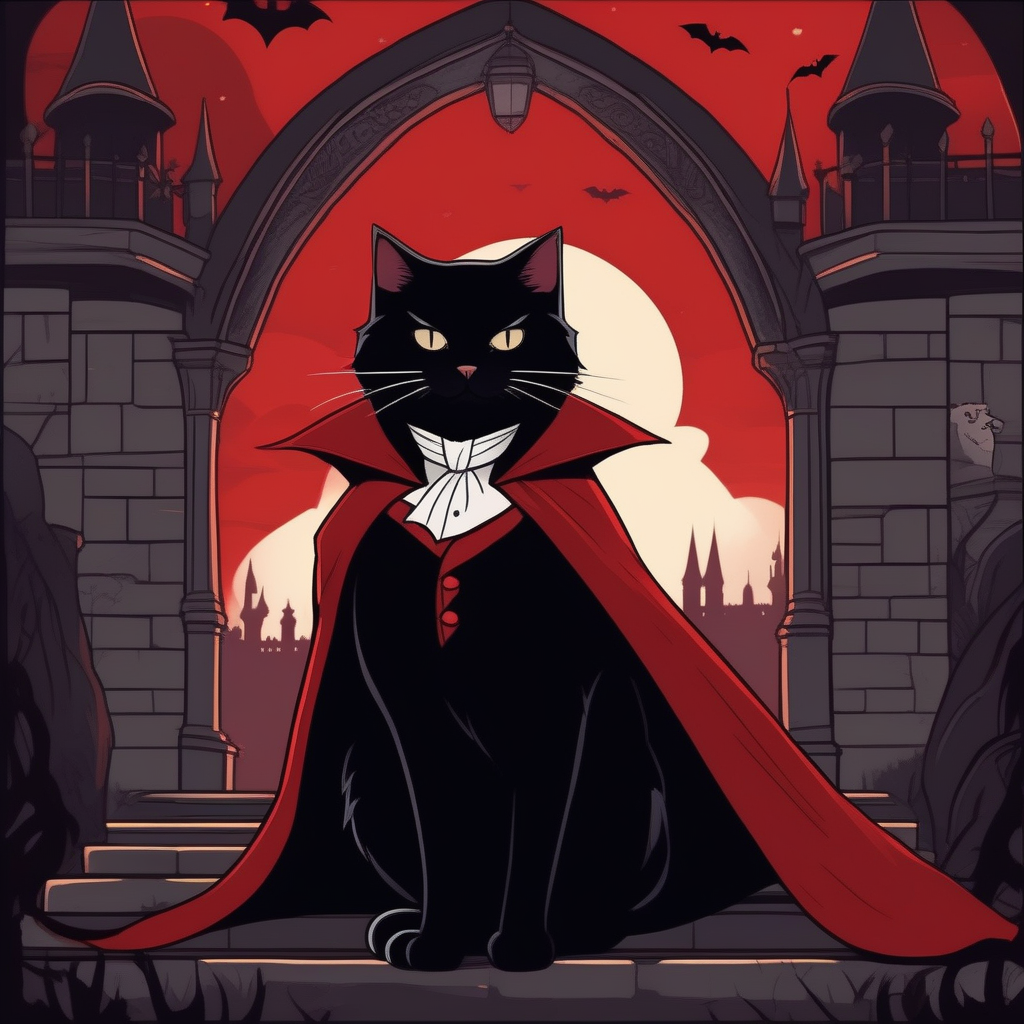} 
\caption{SDXL}
\label{fig:1subim1}
\end{subfigure}
\begin{subfigure}{0.33\textwidth}
\includegraphics[width=0.9\linewidth, height=4cm]{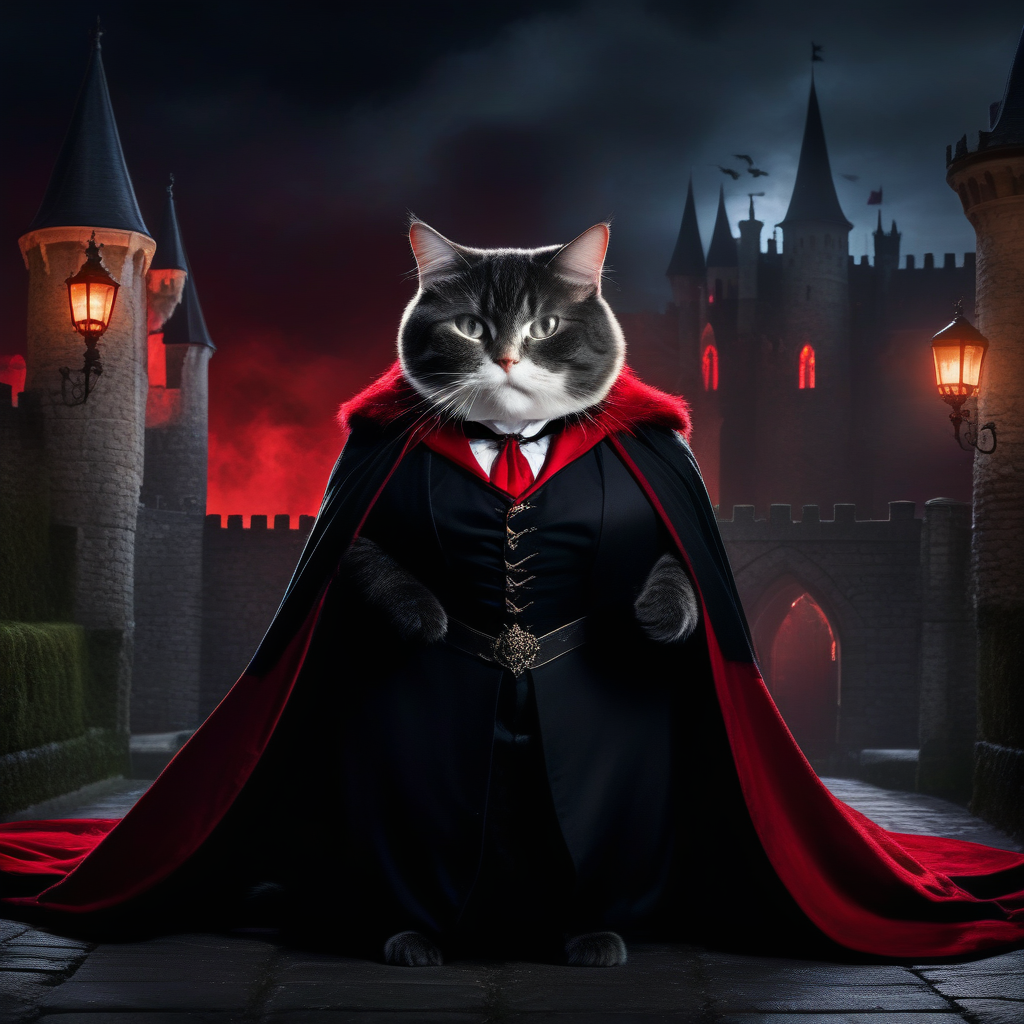}
\caption{SSD-1B}
\label{fig:6subim2}
\end{subfigure}
\begin{subfigure}{0.33\textwidth}
\includegraphics[width=0.9\linewidth, height=4cm]{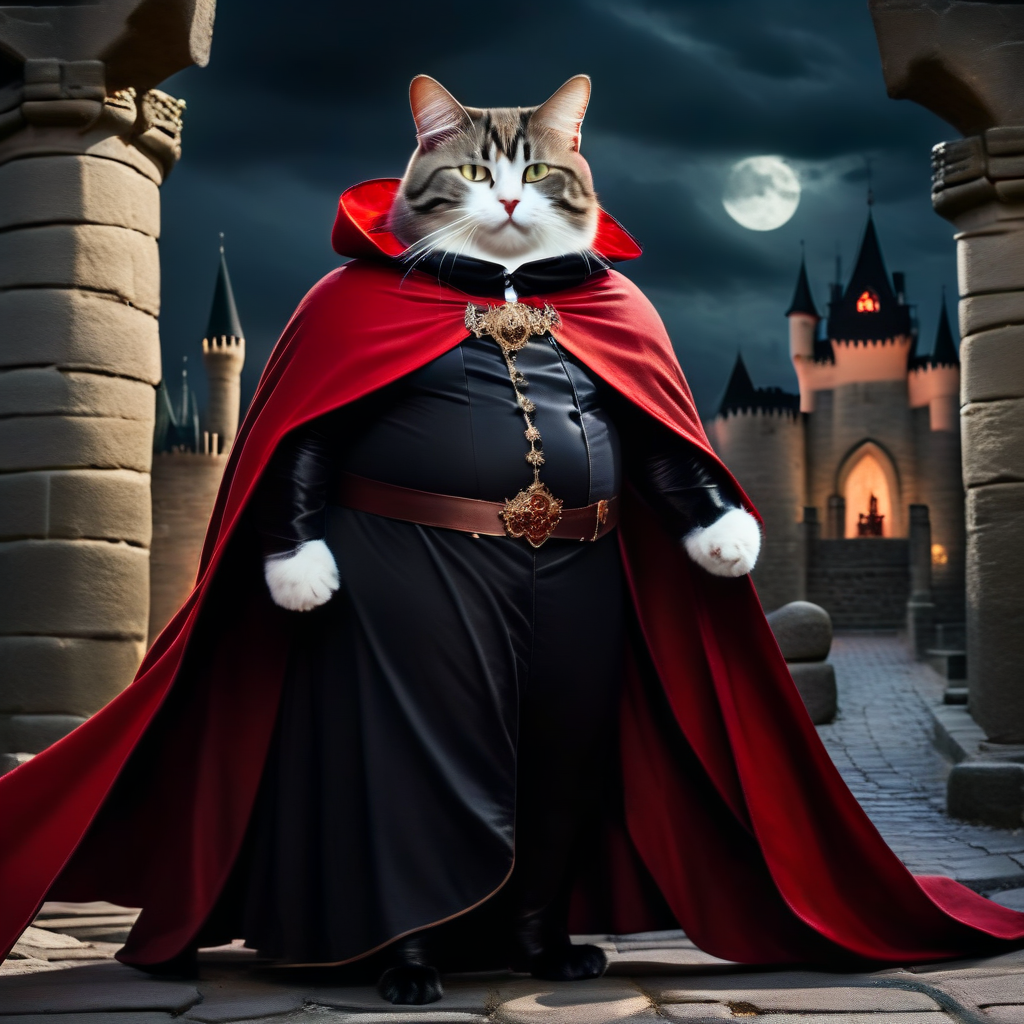}
\caption{Vega}
\label{fig:6subim3}
\end{subfigure}
\caption{"((fatty cat)) dracula, Victorian style, dracula-inspired, long red-black cloak, fangs, castle, in motion, furry paws, action-packed background, dark theme, glow"}
\label{fig:image5}
\end{figure}

We present two distilled versions of Stable Diffusion XL, Segmind Stable Diffusion(SSD-1B) and Segmind Vega, which closely mimic the outputs of the base model as shown in the Figure \ref{fig:image0}, \ref{fig:image1}, \ref{fig:image2}, \ref{fig:image3}, \ref{fig:image4} and \ref{fig:image5}.  All images are generated with the DDPM Scheduler, 25 inference steps and Guidance Scale set to 9.

We report up to 60\% speedup with SSD-1B and up to 100\% speedup with Segmind-Vega. The detailed metrics taken on an A100 at 25 steps with DDPM Scheduler at guidance scale 9 and batch size 1, are reported in Table \ref{tbl:metrics}. 
\begin{table}[!h]
\begin{center}
\begin{tabular}{|c|c|c|}

        \hline
        \textbf{Model} & \textbf{Inference Time (s) ($\downarrow$)} & \textbf{Iteration/s ($\uparrow$) } \\
\hline
SD1.5 \tablefootnote{Inference Times reported at resolution 768 * 768} & 1.699 & 16.79 \\
\hline
SDXL & 3.135 & 8.80 \\
\hline
SSD-1B & 2.169 & 13.37 \\
\hline
Vega &  1.616 & 18.95 \\
\hline 
    \end{tabular}
    \vspace{5mm}
    \caption{Benchmarking inference latency}\label{tbl:metrics}
    \end{center}
    \end{table}
\FloatBarrier

\FloatBarrier
\subsection{Quality Study}

PlaygroundAI\footnote{\url{https://playgroundai.com/}}, a generative AI startup, conducted an extensive blind human preference study encompassing 1000 images and involving 1540 unique users to assess the comparative performance of SSD-1B and SDXL. Remarkably, the findings revealed that not only did SSD-1B maintain image quality, but it was also marginally preferred over the larger SDXL model. The comprehensive details of this study are presented in Table \ref{tbl:pref}.

    \begin{table}[!h]
    \begin{center}
    \begin{tabular}{|c|c|c|}

        \hline
        \textbf{Model} & \textbf{Pairs Won ($\uparrow$)} & \textbf{Percentage Pairs Won ($\uparrow$)} \\
        \hline
         SSD-1B & 528 & 52.8 \\ 
         \hline
         SDXL & 472 & 47.2 \\
         \hline
    \end{tabular}
    \vspace{5mm}
    \caption{Human preference study}\label{tbl:pref}
    \end{center}
    \end{table}
\FloatBarrier
\vspace{-7mm}
The table illustrates the outcomes of the study, with SSD-1B securing victory in 52.8\% of the image pairs, whereas SDXL, although commendable, trailed slightly with 47.2\%. These results not only underscore the noteworthy quality preservation of SSD-1B but also highlight its perceptible preference among the diverse user cohort involved in the blind study.

\section{Conclusion}

We show that distillation of large models like SDXL via using knowledge distillation using multiple models as teachers and using feature losses can quickly converge to give similar quality outputs as the base model despite having a significantly smaller student model. Some of its limitations include but are not limited to Text, Hands and Full Body shots.

Our work also highlights the importance of choice of dataset and teacher model as it can tremendously help boost the final model's quality. We show that progressively distilling can reduce total training time significantly. In the future this technique cnn be further explored on other large models such as LLMs, MLMs etc,

\bibliographystyle{iclr2021_conference}

\end{document}